# FOR-instance: a UAV laser scanning benchmark dataset for semantic and instance segmentation of individual trees

Stefano Puliti [1], Grant Pearse [2], Peter Surový [3], Luke Wallace [4], Markus Hollaus [5], Maciej Wielgosz [1], Rasmus Astrup [1]

[1] Norwegian Institute for Bioeconomy Research (NIBIO), Division of Forest and Forest Resources, National Forest Inventory. Høgskoleveien 8, 1433 Ås, Norway (stefano.puliti@nibio.no; rasmus.astrup@nibio.no)

[2] Scion, 49 Sala Street, Private Bag 3020, Rotorua 3046, New Zealand (grant.Pearse@scionresearch.co).

[3] Faculty of Forestry and Wood Sciences, Czech University of Life Sciences, Prague, Kamýcká 129, 165 00 Praha, Czech Republic (surovy@fld.czu.cz)

[4] School of Geography, Planning and Spatial Sciences, University of Tasmania, Hobart, Australia (luke.wallace@utas.edu.au)

[5] TU Wien, Department of Geodesy and Geoinformation, E120-07, Wiedner Hauptstraße 8, 1040 Vienna, Austria (markus.Hollaus@geo.tuwien.ac.at)

## Abstract

The FOR-instance dataset (available at https://doi.org/10.5281/zenodo.8287792) addresses the challenge of accurate individual tree segmentation from laser scanning data, crucial for understanding forest ecosystems and sustainable management. Despite the growing need for detailed tree data, automating segmentation and tracking scientific progress remains difficult. Existing methodologies often overfit small datasets and lack comparability, limiting their applicability. Amid the progress triggered by the emergence of deep learning methodologies, standardized benchmarking assumes paramount importance in these research domains. This data paper introduces a benchmarking dataset for dense airborne laser scanning data, aimed at advancing instance and semantic segmentation techniques and promoting progress in 3D forest scene segmentation. The FOR-instance dataset comprises five curated and ML-ready UAV-based laser scanning data collections from diverse global locations, representing various forest types. The laser scanning data were manually annotated into individual trees (instances) and different semantic classes (e.g. stem, woody branches, live branches, terrain, low vegetation). The dataset is divided into development and test subsets, enabling method advancement and evaluation, with specific guidelines for utilization. It supports instance and semantic segmentation, offering adaptability to deep learning frameworks and diverse segmentation strategies, while the inclusion of diameter at breast height data expands its utility to the measurement of a classic tree variable. In conclusion, the FOR-instance dataset contributes to filling a gap in the 3D forest research, enhancing the development and benchmarking of segmentation algorithms for dense airborne laser scanning data.

# Introduction

Accurate individual tree segmentation from laser scanning data represents one of the fundamental challenges for utilization of airborne laser scanning in the context of forest ecosystem characterization and sustainable management of forests. Information on individual trees and their properties is becoming increasingly required to manage and study forests in order to ensure the provision of a broad range of products (timber and non-timber products) and services (e.g. biodiversity, water, carbon storage). For example, adopting continuous cover or close-to-nature forestry practices shifts the management unit from the stand to the individual tree and requires a greater granularity in forest information compared to current forest resource maps. Dense laser scanning data acquired from terrestrial or aerial platforms provide an unparalleled 3D representation of individual trees and forest canopies and thus have emerged as promising data source for direct measurement of single-tree information. However, despite the wealth of literature on the use of dense laser scanning data, the automation of the segmentation of individual trees (i.e. instance segmentation) and their respective structures (i.e. semantic segmentation) from dense forest point clouds remains a significant challenge that restricts the full utilization of these data.

In addition to the intrinsic challenge in separating single tree crowns that, due to their nature, can overlap with their neighbors, one of the main limitations in developing individual-tree instance segmentation methods has been the lack of publicly available, curated, and machine learning (ML) ready benchmarking datasets for boosting and tracking our research progress (Lines et al. 2022). As forest vary considerably, in terms of tree shapes, structures and complexity it is difficult to evaluate the performance of various segmentation approaches if they are not compared on the same independent dataset. While we are currently seeing some of the first open datasets with annotated individual trees, such as the Weiser et al. (2022) and the LAUTx datasets (Tockner et al. 2022), to date, most studies proposed methodologies tailored to relatively small datasets, generally leading to using overfitted models which can hardly transfer to new data. Furthermore, the absence of open benchmark datasets has led to the validation of methods primarily with internal datasets, potentially inflating the performance metrics and inhibiting the direct comparison of different approaches using fully independent datasets. Moreover, with the rapid increase in the availability of deep learning instance and semantic segmentation models for forest point clouds (e.g. Krisanski et al. 2021; Wielgosz et al. 2023b; Windrim and Bryson 2020) it is imperative to define common benchmarking schemes to ensure a transparent and efficient algorithm development. Furthermore, currently the availability of dense laser scanning data is geographically skewed and is generally sparse or non-existent in many developing countries. This highlights a clear need for bridging this data divide and promoting data accessibility and inclusivity across the globe to build a broader and stronger scientific community ultimately.

In this data paper, we present a benchmarking dataset for semantic and instance segmentation of dense airborne laser scanning data captured from UAV or low flying helicopters. The function of the dataset is to provide development and testing data across a wide range of forest structures and support the development of improved segmentation approaches for forest ecosystems. t. By providing pre-defined development and evaluation datasets, we enable researchers to develop novel methods and effectively track the research progress for dense point cloud segmentation of forested 3D scenes.

# Data description

FOR-instance is a curated and ML-ready dataset of five UAV-LS collections acquired in Norway, the Czech Republic, Austria, New Zealand, and Australia (see Figure 1). Different partners kindly shared the UAV-LS data as part of an open-data effort to develop a common and open benchmarking infrastructure for more effective algorithm development and progress tracking.

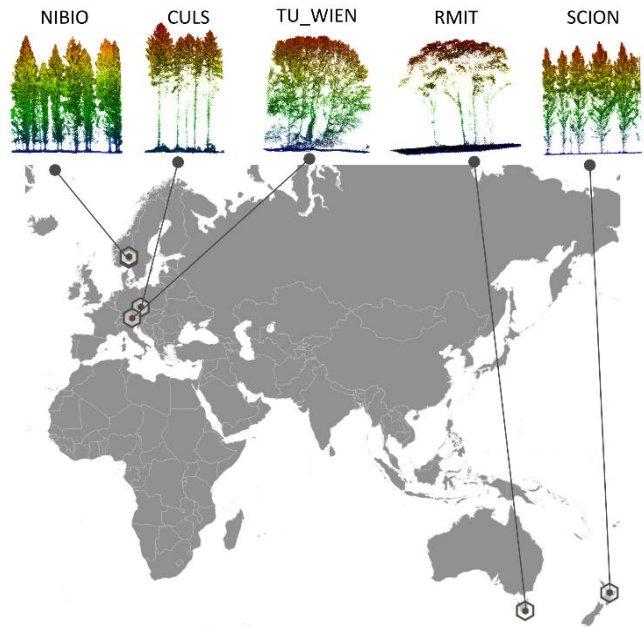

**Figure 1**. A geographical overview of the datasets included in the FOR-instance data.

The different collections represented previously existing datasets, and as such, they differed in sampling designs, forest types/management, sensors, and data collection patterns resulting in varying degrees of 3D scene coverage (plots or wall-to-wall) and quality (varying point density). Table 1 summarizes the main characteristics of the different collections.

**Table 1**. Summary of the key characteristics of the FOR-instance benchmark data.

| Collection | Country | Reference | n plots | n trees | Total annotated area (ha) | Forest type | Sensor | Average point density (pts m$^{-2}$) |
|---|---|---|---|---|---|---|---|---|
| **NIBIO** | Norway | Puliti et al. (2022) | 20 | 575 | 1.21 | coniferous dominated boreal forest | Riegl miniVUX-1 UAV | 9529 |
| **CULS** | Czech Republic | Kuželka et al. (2020) | 3 | 47 | 0.33 | coniferous dominated temperate forest | Riegl VUX-1 UAV | 2585 |
| **TU_WIEN** | Austria | Wieser et al. (2017) | 1 | 150 | 0.55 | deciduous dominated alluvial forest | Riegl VUX-1 UAV | 1717 |
| **RMIT** | Australia | Unpublished | 1 | 223 | 0.37 | Native dry sclerophyll eucalypt forest | Riegl MiniVUX-1 UAV | 498 |
| **SCION** | New Zealand | Unpublished | 5 | 135 | 0.33 | Non-native pure coniferous temperate forest | Riegl MiniVUX-1 UAV | 4576 |

## Sampling designs

Due to the varying nature of the data acquisitions and the limited annotation possibilities, the data coverage varied between the collections. It was available at plot level (NIBIO, CULS, SCION) or wall-to-wall for larger contiguous areas (RMIT, TU_WIEN).

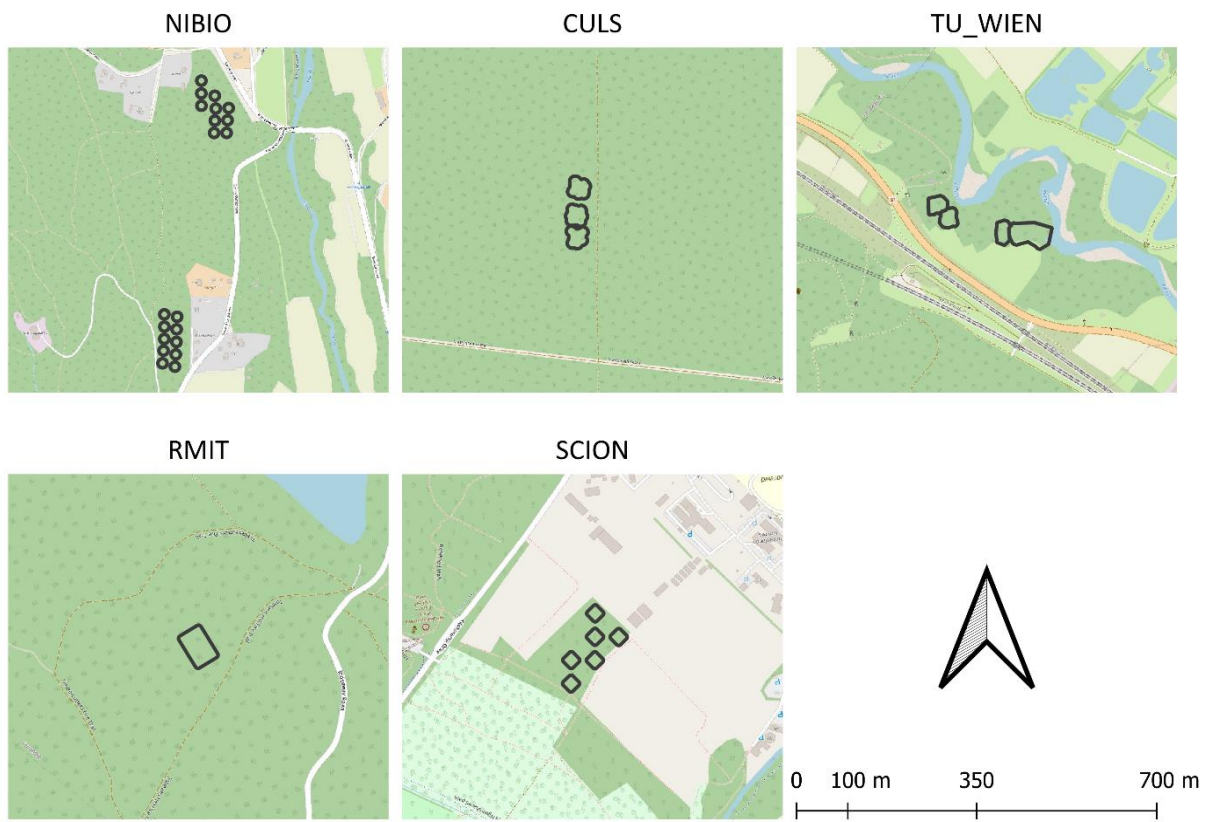

**Figure 2**. Variations in terms of sampling designs in the FOR-instance collections

## Forest types

FOR-instance includes the following five forest types:

I. **Coniferous dominated boreal forest (42% of the annotated trees)**: these are found in Norway in the NIBIO collection and are composed of mature forests dominated by Norway spruce (*Picea abies*) and with few Scots pine (*Pinus Sylvestris*) and deciduous species (mainly *Betula sp.*). In specific, the NIBIO collection is mainly dominated by Norway spruce. Further details on the NIBIO forest structures can be found in the study by Puliti et al. (2022).

II. **Coniferous dominated temperate forest (5% of the trees)**: these forests correspond to three plots in the CULS collection and consist of a productive even-aged Scots pine forest stand found in sandy areas of the Czech north of Prague, that are generally more productive than the mire forest in boreal conditions. Further details regarding forest characteristics can be found in Kuželka et al. (2020).

III. **Deciduous alluvial forests (29% of the annotated trees)**: this forest type is found in the TU_WIEN collection and consists of a Natura 2000 site (Niederösterreichische Alpenvorlandflüsse; Area code: AT1219000) characterized by complex vegetation structures (woody debris, lying and standing deadwood, dense understory) a large diversity of tree species and considerable variation in canopy layering and tree size (DBH up to 1.3 m). Further details on the specific forest characteristics can be found in Wieser et al. (2017).

IV. **Native dry sclerophyll eucalypt forest (11% of the annotated trees)**: This forest type is found in the RMIT collection with a dominant canopy species of White Peppermint (*Eucalyptus pulchella).* Trees of this species are found within the plot at a range of ages and heights (from 5 m to 17 m). The study site was subject to planned burning 3 years before the drone laser scanning acquisition. The understory at time of capture consisted of low shrubs (up to 2 m in height) and native grasses.

V. **Coniferous non-native temperate (13% of the annotated trees)**: This type is found in the SCION collection and consists of pure radiata pine (*Pinus radiata D. Don*) of 18-20 years of age and is representative of a typical productive plantation forest (approx. 930 trees per hectare) in New Zealand. The forest was planted on a 2 m × 5 m grid and remained un-thinned, and no silvicultural interventions except for regular mowing of the blackberry understory between the rows. The vertical forest structure is single-layered, and due to the branchiness of radiata pine, the canopy is composed of a dense layer of intertwined and overlapping crowns.

While the tree species information was unavailable for all collections and thus not published in FOR-instance data, most trees were coniferous (60 %) and primarily found in productive forests. Regarding canopy layering, approximately 89 % of trees of the entire dataset have at least a portion of the crown visible from above the canopy, whereas the remaining 11% of the trees are growing in the understory and do not have the crown visible from above.

## DBH in-situ measurement

Field data campaigns were conducted in each of the FOR-instance collections to measure the diameter at breast height (DBH; cm), a typical variable used in forest inventories. The DBH measurements were conducted by measuring the DBH using a measuring caliper for all of the trees within the sampled areas with a diameter at breast height (DBH)≥ 5 – 7 cm. The overall DBH distribution by collection (see Figure 3) shows a broad range of DBH (5 – 133 cm); on average, the DBH was 23.6 cm.

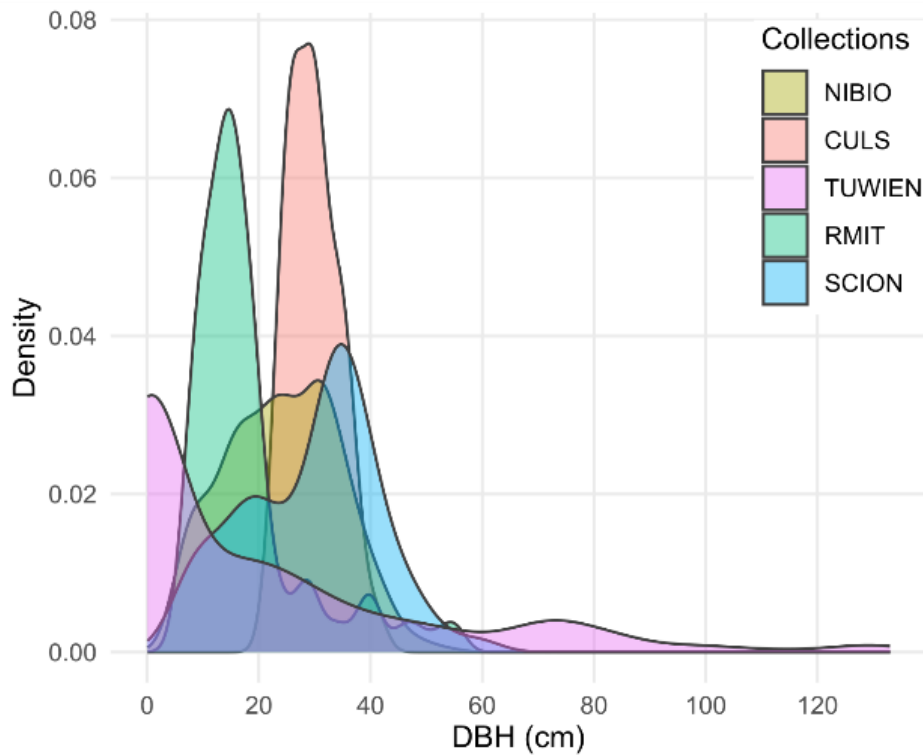

**Figure 3**. Diameter at breast height (DBH) density distribution for each FOR-instance collection.

The DBH range and variation, and thus the forest complexity, was the largest for the TU_WIEN collection, characterized by a "J" shaped distribution composed of many small and few large trees. On the other hand, the distribution of the NIBIO, CULS, and to a certain degree, the RMIT collections were bell-shaped, with the CULS collection showing the narrowest variation in DBH, reflecting a very simple forest structure.

### Drone laser scanning acquisition

The UAV-LS data were acquired with two sensors (see Table 1): the Riegl VUX-1UAV and the Riegl MiniVUX-1 UAV. In regard to the variation in data acquisition protocols, the FOR-instance dataset includes two flight data collection patterns, namely:

- **Parallel**: This consists of flying parallel flight lines
- **Double grid**: the flight lines are planned in two perpendicular directions (e.g. north-south and east-west), forming a square grid, and an additional grid is flown by rotating the original grid by 45˚ (e.g. north east - south west and north east - south west). Such a pattern was flown to reduce occlusions and allow for a more complete and denser dataset.

In addition to flight line patterns, the FOR-instance collections differed regarding other flight and sensor parameters (see Table 2).

**Table 2**. Summary of the key data collection parameters in the different FOR-instance collections.

| Collection | Leaf condition | Flight pattern | Flight altitude (m above ground) | Flight speed (m sec$^{-1}$) | Flightline spacing (m) | Pulse repetition rate (kHz) | Maximum number of returns |
|---|---|---|---|---|---|---|---|
| **NIBIO** | Leaf-on | Double grid | 60 | 5 | 6 | 100 | 5 |
| **CULS** | Leaf-on | Double grid | 80 | 6 | 50 | 550 | 5 |
| **TU_WIEN** | Leaf-off | Parallel | 50 | 8 | 40 | 350 | 3 |
| **RMIT** | Leaf-on | Parallel | 40 | 5 | 10 | 100 | 5 |
| **SCION** | Leaf-on | Double grid | 50 | 5 | 15 | 100 | ≤ 5 |

## Point cloud annotation

The data were fully annotated into single trees (i.e. instance annotation) and different parts of the trees and the forest (i.e. semantic annotation) using CloudCompare (CloudCompare, V 2.12.4) over approximately six months by a team of two annotators and later reviewed. The annotation consisted of a two-step procedure (see Figure 4) consisting of instance and semantic annotation.

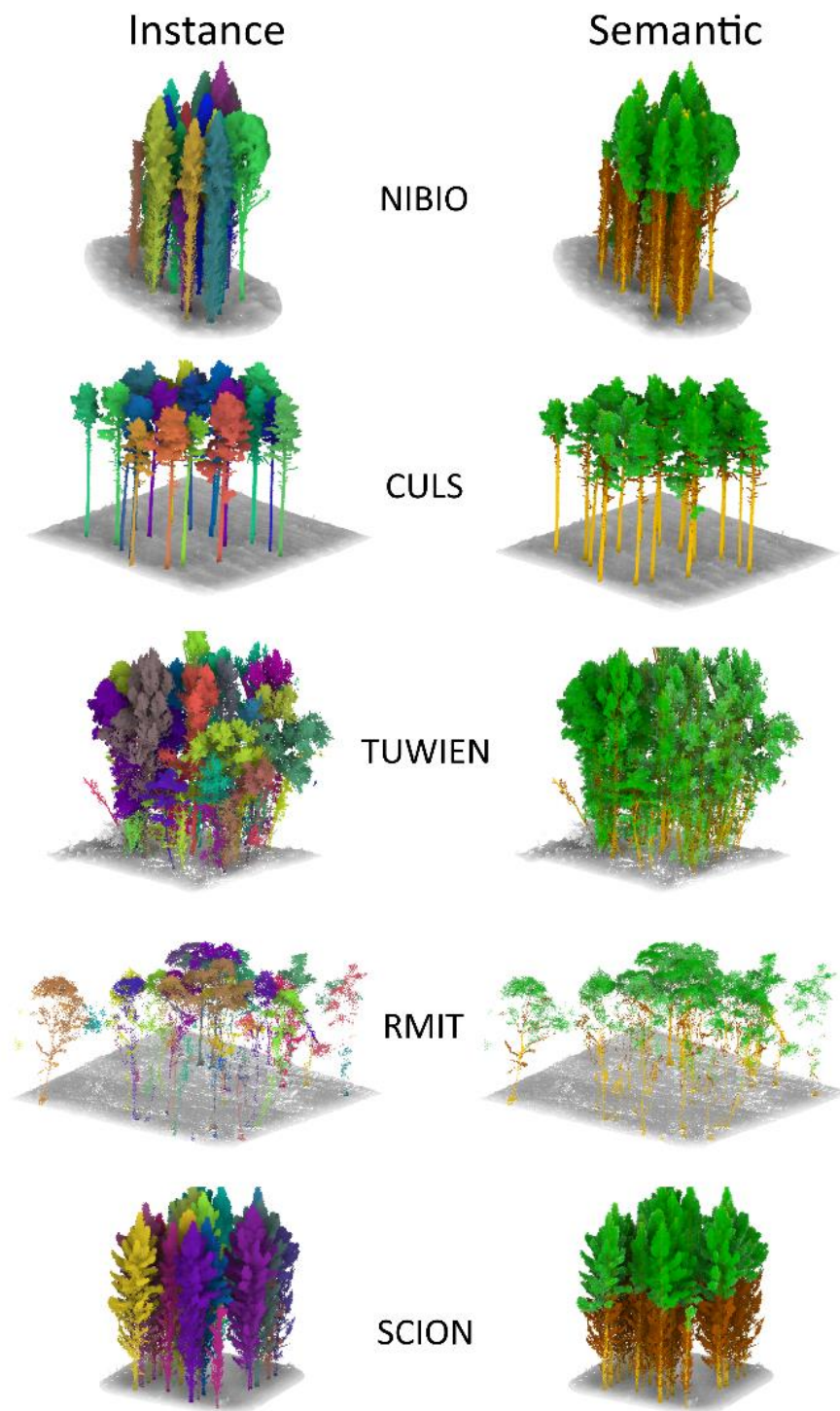

**Figure 4**. Examples of instance and semantic annotations for samples of the different FOR-instance data collections.

Instance annotation

In this step, the point clouds corresponding to each tree within the plot boundary were separated from the rest of the trees, and a unique tree identifier within each collection was assigned to all of the points representing the tree of interest. The individual tree annotation was done by iteratively removing points (i.e. using the segmentation tool in CloudCompare) from neighboring trees and by making sure as far as practicable to separate also intertwined and overlapping crowns (see Figure 5).

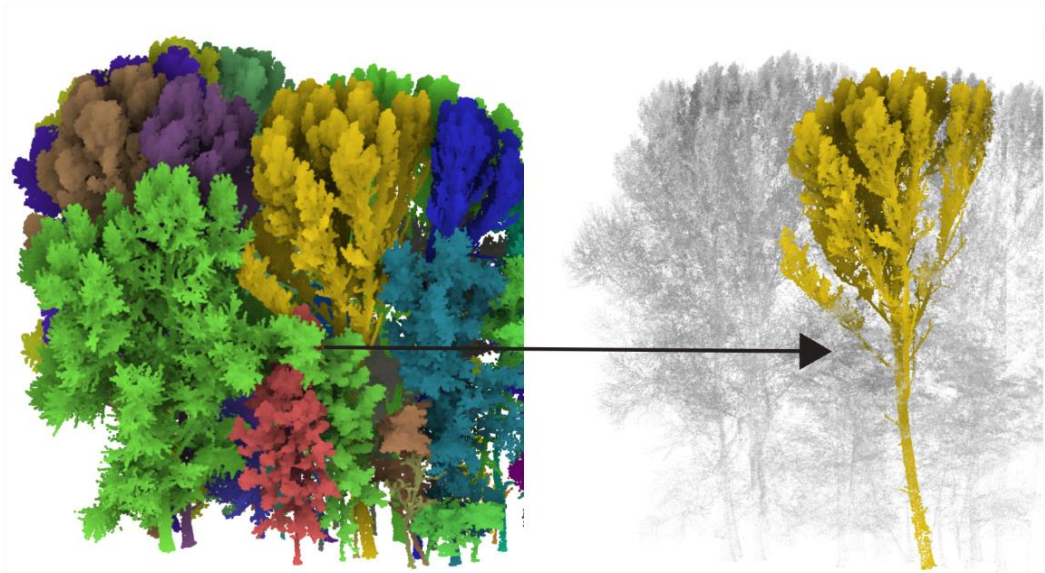

**Figure 5**. Example of instance annotation output from a plot (TU_WIEN collection) characterized by a complex forest structure where an individual tree is separated from the rest of the point cloud with a detail (i.e. left pane) showing the quality of the instance annotation at the crown- and branch-level.

The tree unique identifier was stored as an additional *treeID* column in the .las files. The annotation was performed for all trees where the stem was visible in the point cloud. Hence, small trees in the understory were assigned to a low vegetation class (see step 2 semantic annotation). Trees outside the plot boundaries were not annotated but retained and assigned to the out-points class (see step 2 semantic annotation). In addition to any other point that did not represent a tree (e.g. ground), these points were assigned a *treeID* = 0.

Semantic annotation

Once all of the individual trees were identified, we further annotated each tree point cloud into the following semantic classes (see Figure 5 and Table 3 for classification codes):

- **Stem (5.1% of annotated points)**: Includes points corresponding to the central tree stem axis. Due to the large occlusions in the higher part of the trees, compounded with the smaller dimension of the stem, the quality of the separation of the stem was poorer in the higher parts of the tree compared to the base of the tree. In addition, parts of the live branches were annotated as stems for some of the collections (e.g. NIBIO) due to some over-simplification on the tree tops.
- **Woody branches (4.66% of annotated points)**: This class includes branches composed of woody material without any leaves or needles attached. While this class is far from a perfect wood-leaf separation, it would still allow to separate the live crown from low branches and thus enable the measurement of the live crown base and related metrics. The separation between these "non-green" branches and live branches was most challenging for coniferous trees due to the small diameter of the branches and the presence of a thick layer of needles causing substantial occlusions. Further challenges in this class annotation were due to the significant time consumption required for annotating fine branches, resulting in some overlap (i.e., noise) between the branches and the live branch classes. These limitations should be considered when developing or evaluating semantic segmentation models.
- **Live branches (63.8% of annotated points)**: This class represents the most significant portion of the annotated data and includes branches with leaves or needles attached. As mentioned above, the separation between these and the woody branch class was challenging, and thus users should be careful when using these semantic labels as they do not reflect a finely detailed leaf-wood separation but rather provide information on a general crown distribution of the live crown.

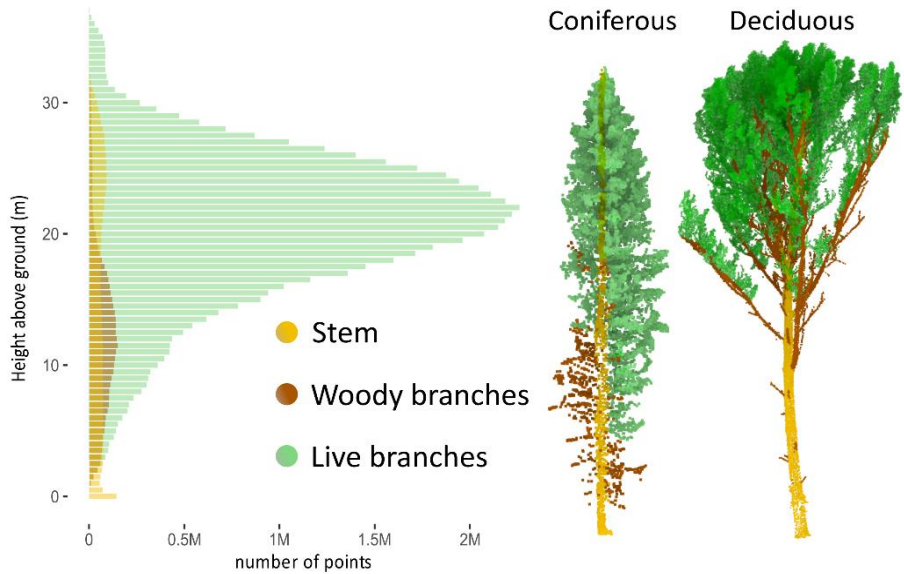

**Figure 5**. Distribution of the points across tree semantic classes through the vertical canopy profile and detail on two example trees in the FOR-instance data to illustrate the differences in the semantic classes for coniferous and deciduous trees.

Furthermore, additional non-tree-related semantic classes were annotated for a complete panoptic annotation of the entire 3D scene. These consisted of:

- **Unclassified (0.01% of the annotated points)**: these represented a tiny portion of the points that could not be annotated, i.e. assigned to any of the trees, but that was retained only for completeness of the 3D scene. These points can be masked from the point clouds for method development and benchmarking with negligible effects on the results.
- **Low-vegetation (17.42% of the annotated points)**: these points were derived by a combination of i) removal of individual trees and ii) iterative terrain removal using the cloth simulation filter (CSF) implemented in CloudCompare. Hence this class includes small trees, understory vegetation, bushes, and herbaceous vegetation that could not be assigned a unique tree identifier. It is important to note that a relatively aggressive ground removal was applied; thus, this class can partly overlap with the following terrain class. Users can optionally merge these two classes and apply a preferred ground classification algorithm.
- **Terrain (8.96% of the annotated points)**: as mentioned for the low-vegetation class, the ground was derived from an automated and iterative application of the CSF algorithm until most vegetation was removed while at the same time ensuring a relatively homogeneous coverage (i.e. lack of wide data gaps). This class should not include any green vegetation.
- **Outpoints (42% of the total points)**: correspond to trees or other vegetation outside the plot boundary. While not annotated, these points were retained for completeness of the 3D scene and should be used when performing inference, but as they do not have a label should not impact the evaluation metrics.

The different semantic classes were stored in the classification column in the .las files according to the codes described in Table 3.

**Table 3**. Description of the semantic classes for annotating the FOR-instance dataset with the corresponding Classification code.

| Classification Code | Class name |
|---|---|
| 0 | Unclassified |
| 1 | Low-vegetation |
| 2 | Terrain |
| 3 | Out-points |
| 4 | Stem |
| 5 | Live branches |
| 6 | Woody branches |

### Known issues and limitations due to the annotation.

While annotation was generally feasible, accurately interpreting forest point cloud scenes remains challenging, largely depending on the point cloud's density and quality as well as the tree size and visibility (i.e. occlusions). Thus, the proposed annotation represents the best effort to provide a high-quality instance and semantic segmentation dataset. However, some issues are present in the datasets and include:

- Challenges in leaf / wood separation
- Separation between ground and low vegetation
- Presence of unclassified points

## Data split

The annotated collections were then split into development (i.e. dev) and test sub-sets in order to create two separate sets of data for the following scopes:

- **Dev data**: these data can be used for developing new methods. The dev data can be further split into training and validation by the user to create a validation dataset for hyperparameter tuning, overfitting detection, and model selection.
- **Test data**: these data aim to be used for the final evaluation of a method (i.e., either pre-existing or developed using the training data). These datasets should strictly not be used in any step of model training, and to avoid any information leak, the user should limit the number of times the test data is used.

For this purpose, approximately 70% of the annotated area was selected for development and 30% for testing. For the collections where the sampling design was in the form of fixed-area plots (i.e. NIBIO, NIBIO2, CULS, and SCION), the split was performed by selecting a random sample of 70% of the number of plots for development, and the remaining 30% for testing. In the case of the collections that were annotated wall-to-wall (i.e. TU_WIEN and RMIT), a randomly selected area of approximately 30% of the total was retained for testing, while the remaining 70% was used for development. Within the FOR-instance data, the data split is provided in a table (metadata_split.csv) where each file name is labeled as *dev* or *test* in the *split* column.

# Data usage

One of the driving principles behind the creation of the FOR-instance data was the idea of providing the international scientific community with open access to a fully annotated dataset covering a range of forest types to accelerate the research progress in point cloud segmentation tasks. Specifically, the FOR-instance dataset was conceived to provide two main uses, namely method development and benchmarking to address both i) the lack of open ML-ready datasets for developing new instance and semantic segmentation methods and 2) the lack of common benchmarking infrastructure for efficient progress tracking respectively.

While the primary focus was on the instance segmentation of trees in dense 3D point clouds, the FOR-instance data may be used for several other applications. In the following sections, we outline some potential usage applications and describe best practices for benchmarking.

## Method development

The use of the FOR-instance data for developing new methods is done by using the dev data (i.e. see Data split section) as input and output. In its most basic form, the development of new machine- or deep-learning methods using the FOR-instance, consists of:

1. **Splitting the dev data into training and validation** is optional, but helpful to set aside some validation data during the method development for both hyperparameter tuning and best model selection.
2. **Train a model using the training data**: the training data is used to estimate model parameters. The format of the data with associated point labels to train a model;
3. **Best model selection using the validation**: using the validation data during the model training helps detect overfitting and select the best model. It is important to remark that the validation data should not be used for reporting accuracy metrics, which should only be computed using the test data (see Benchmarking section) once the final best model is selected using the validation data.

## Input data formats

In terms of data format, given the native point cloud format of the data, the use of the raw point cloud with associated point labels in 3D deep learning frameworks is the most intuitive. However, given the performance of image-based deep learning methods, it is also possible to transform the input point cloud data into 2D projections of the point clouds or images (see Figure 6). As an example, following this idea, Straker et al. (in press) converted the FOR-instance point clouds into rasterized canopy height models (CHM) to study the efficiency of using a data-driven YOLOv5 v7.0 image segmentation model (Jocher et al. 2022) compared to traditional CHM algorithmic approaches (e.g. tree top detection, watershed segmentation).

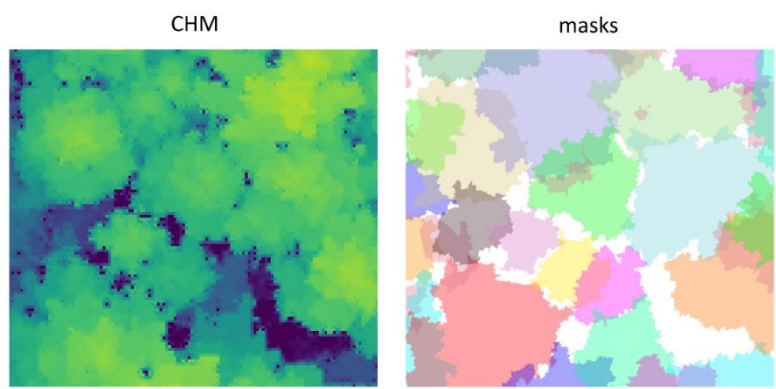

**Figure 6**. Examples of how the FOR-instance data can be prepared to train an image-based CNN, for instance segmentation using canopy height models (CHM) and tree crown masks.

### Semantic segmentation

The classification field in the .las files can be used as a reference label for training semantic segmentation models. It is important to highlight that, thanks to the granularity of the semantic labels, in FOR-instance, it is possible to aggregate some of the classes while maintaining a biological meaning in the semantics. Below are some possible and interesting examples of class aggregation of the tree semantic classes:

- **Woody branches + live branches**: these two classes can be merged to create a crown class, thus providing a stem-crown separation equivalent to the semantic annotation proposed by Wielgosz et al. (2023a). Such a merger can be helpful, especially in coniferous forests where trees are characterized by a single main stem holding the most significant portion of the tree biomass.
- **Woody branches + stem**: this merging allows to merge into a single class the tree's woody components, thus providing the commonly used wood-leaf/needle separation. It is important to note that the leaf/needle class also includes some branches, especially in coniferous forests.

### Instance segmentation

The *treeID* field in the .las files can be used as reference labels to train instance segmentation models in 3D and 2D deep learning frameworks. In addition to the available sensor data, i.e. x, y, z, intensity, return number, number of returns, and scan angle, users can further append new features such as geometrical features derived from normal vectors.

Concerning the plot edges, users can define whether the outer points (i.e. points without *treeID*) are included in the training. In the former case, the training data will consist of the entire forest, where some tree instances are annotated while others are not. This creates a somewhat realistic forest but also creates the issue that some instances are not labeled in the training data. In the latter case, the model will consider the training data as isolated forest patches, which eliminates the issue of having unlabeled instances in the training data but also oversimplifies the forest structure.

### DBH

The use of the DBH information can be twofold:

- **Calibrating/developing DBH estimation method**s might include circle or cylinder fitting and modeling the DBH from tree predictor variables.

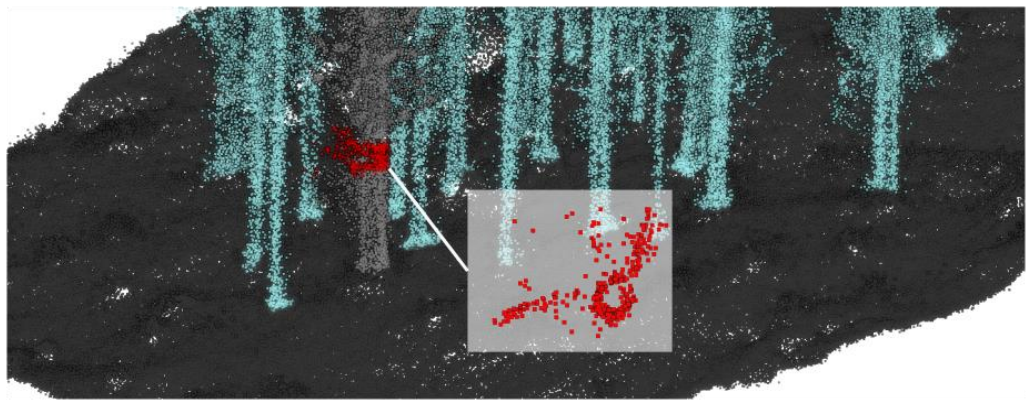

**Figure 7**. Example of tree slice at breast height that can be used for direct DBH measurements from the point cloud.

- **For subsetting trees by their size**: this might be helpful to, for example, set a DBH threshold for removing small trees from the training or selecting balanced DBH distribution in the training dataset.

## Benchmarking

The following sections describe the best practices for benchmarking semantic instance segmentation and DBH measurements using the FOR-instance test data (see Data split section).

### Semantic segmentation

Given the 3D nature of the data, the basis for evaluating any semantic segmentation should be the point cloud. Given the available reference class label and a predicted label, the user can construct a confusion matrix (see Figure 8) from which true positives (TP), true negatives (TN), false positives (FP), and false negatives (FN) are obtained. Given a number $n$ of reference data points and a number $c$ of semantic classes, the overall accuracy can be computed according to:

$$Overall\ accuracy\ (OA) = \frac{\sum_{i=1}^{c} TP_c}{n} \quad (1)$$

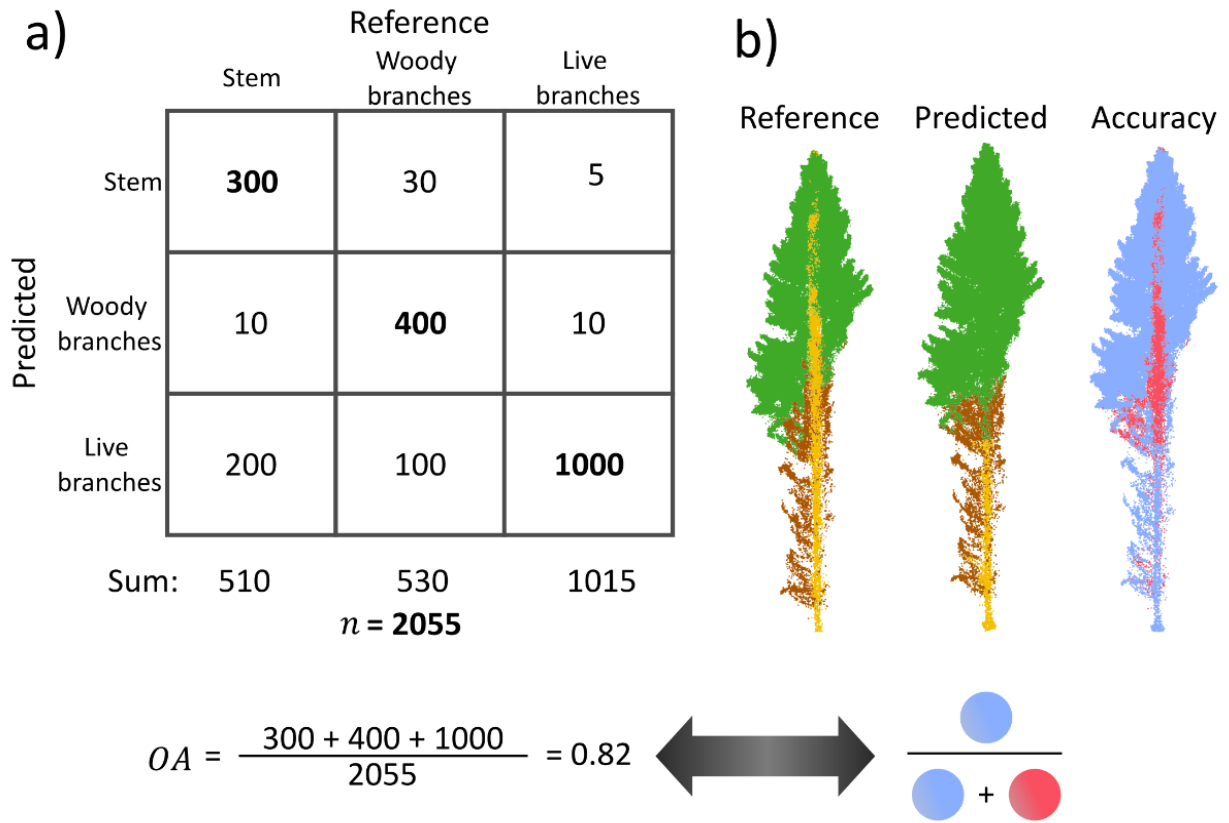

**Figure 8**. Example of a confusion matrix (note that values are fictitious) for the semantic segmentation of a single tree (a), visualization of how the confusion matrix is constructed at the point level (b), and calculation of the overall accuracy ($OA$).

For each of the classes it is then possible to derive the precision (P), recall (R), and F1 score (F1) as measures of the user's and producer's accuracy and a combined assessment of the two computed as follows:

$$Precision\ (P) = \frac{True\ positive\ (TP)}{True\ positive\ (TP) + False\ positives\ (FP)} \quad (2)$$

$$Recall\ (R) = \frac{True\ positive\ (TP)}{True\ positive\ (TP) + False\ negatives\ (FN)} \quad (3)$$

$$F1\ score\ (F1) = 2 \cdot \frac{Precision\ (P) \cdot Recall\ (R)}{Precision\ (P) + Recall\ (R)} \quad (4)$$

In cases such as in the FOR-instance dataset, where there is a strong imbalance for the different classes, the balanced accuracy ($BA$) might provide a better metric than the overall accuracy. The $BA$ is then computed as the average of the recall values for each class.

$$Balanced\ accuracy\ (BA) = \frac{\sum_c recall\ (R)}{c} \quad (5)$$

### Instance Segmentation

The first step in performing any instance segmentation evaluation is to match the point cloud reference and predicted instance IDs. To do this, it is necessary to define whether a prediction is correct. Here we adopt a method Wielgosz et al. (2023) proposed for matching tree instance IDs from two point clouds with reference and predicted instance IDs, respectively.

Given a reference and a predicted point cloud with two separate sets of instances, the tree instances are iteratively matched in descending order from the tallest to the shortest trees by using the following algorithm:

a) Find the tallest tree in the reference data;
b) Find the tree in the predicted instances that have the largest intersection over union (IoU) with the tree selected in the previous step;
c) if the IoU is <0.5: the predicted tree is considered an error, and thus no reference instance ID is available;
d) if the IoU is ≥0.5: the tree is considered a correct match, and assign reference instance ID label to the predicted tree;
e) Add to collection (dictionary) of predicted tree instances with IDs matching the reference instance IDs.

Following the initial matching of reference and predicted instance IDs, the evaluation can be done at multiple levels:

- **Tree detection**: a first elementary evaluation can be done on the tree level to evaluate detection, omission, and commission rates (see Figure 9) based on the following equations:

$$Detection\ rate\ (DR) = \frac{TP}{TP + FN} \quad (6)$$

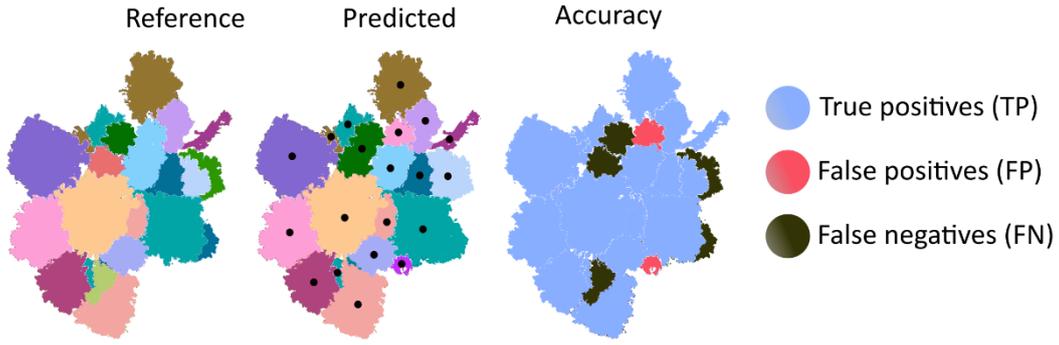

**Figure 9**. Visual example (top-down view of the tree crowns) of how tree detection, omission, and commission rates should be computed at the tree level.

- **Tree segmentation**: Similar to the evaluation of the semantic segmentation, this should be performed at the point cloud level (see Figure 10), and for each tree, one can construct a confusion matrix and derive P, R, and F1 score according to Equations 2, 3, and 4.

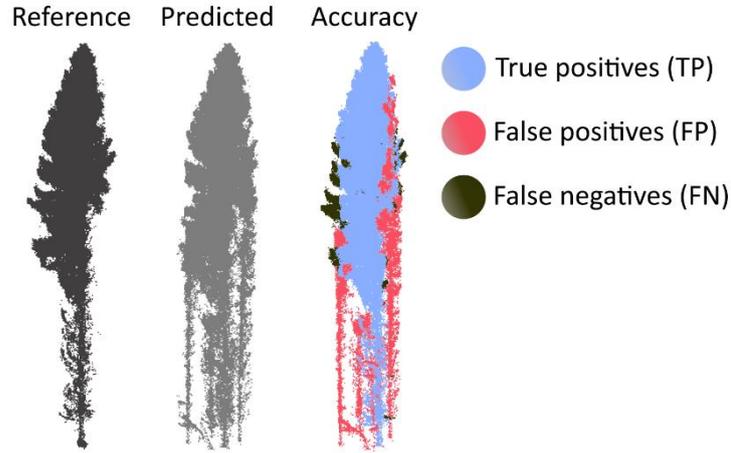

**Figure 10**. Example of a single tree instance segmentation result and the metrics are computed at the point cloud level.

- **Height measurement**: This represents a valuable metric to quantify whether the instance segmentation allows a proper extraction of tree instances on the tree tops. In this case, the root mean square error (RMSE), mean difference (MD), and the respective values expressed as a percentage of the mean (RMSE% and MD%) should be assessed as measures of the random and systematic errors according to:

$$RMSE = \frac{\sqrt{\sum_t (y - \hat{y})^2}}{t} \quad (7)$$

$$MD = \frac{\sum_t (y - \hat{y})}{t} \quad (8)$$

Where $t$ is equal to the total number of trees either in a FOR-instance collection or in the whole dataset, $y$ is the reference height value and $\hat{y}$ is the predicted tree height value in meters above ground (m).

- **Forest segmentation**: This consists of aggregating the tree-wise F1 scores to the level of a plot, collection, or for the entire FOR-instance dataset. This is done by averaging tree-wise F1 scores to the desired area level. There are different weighting coefficients which may be used in order to average F1 scores.

### DBH measurement

As with height measurement, the DBH should be evaluated based on the RMSE, MD, RMSE%, and MD% computed with the Equations 6 and 7 with the only difference that the $y$ and $\hat{y}$ are a reference and predicted DBH (cm).

### Acknowledgements

This work is part of the Center for Research-based Innovation SmartForest: Bringing Industry 4.0 to the Norwegian forest sector (NFR SFI project no. 309671, smartforest.no).